\documentclass[letterpaper, 10 pt, conference]{ieeeconf}  

\IEEEoverridecommandlockouts                              

\usepackage{graphics} 
\usepackage{graphicx} 
\usepackage{amsmath} 
\usepackage{amssymb,bm}  
\usepackage[style=ieee]{biblatex}
\usepackage{lipsum}
\usepackage{color}
\usepackage{balance}

\DeclareSourcemap{
  \maps{
    \map{
      \pertype{article}
      \step[fieldset=url, null]
      \step[fieldset=doi, null]
      \step[fieldset=issn, null]
      \step[fieldset=isbn, null]
      \step[fieldset=note, null]
      \step[fieldset=editor, null]
      \step[fieldset=urldate, null]
      \step[fieldset=file, null]
    }
  }
}
\DeclareSourcemap{
  \maps{
    \map{
      \pertype{inproceedings}
      \step[fieldset=url, null]
      \step[fieldset=doi, null]
      \step[fieldset=issn, null]
      \step[fieldset=isbn, null]
      \step[fieldset=note, null]
      \step[fieldset=editor, null]
      \step[fieldset=urldate, null]
      \step[fieldset=file, null]
    }
  }
}
\DeclareSourcemap{
  \maps{
    \map{
      \pertype{incollection}
      \step[fieldset=url, null]
      \step[fieldset=doi, null]
      \step[fieldset=issn, null]
      \step[fieldset=isbn, null]
      \step[fieldset=note, null]
      \step[fieldset=editor, null]
      \step[fieldset=urldate, null]
      \step[fieldset=file, null]
    }
  }
}
\DeclareSourcemap{
  \maps{
    \map{
      \pertype{misc}
      \step[fieldset=url, null]
      \step[fieldset=doi, null]
      \step[fieldset=issn, null]
      \step[fieldset=isbn, null]
      \step[fieldset=note, null]
      \step[fieldset=editor, null]
      \step[fieldset=urldate, null]
      \step[fieldset=file, null]
    }
  }
}

\title{\LARGE \bf
Validation of Tumbling Robot Dynamics with\\ Posture Manipulation for Closed-Loop Heading Angle Control
}

\author{
Adarsh Salagame$^{1}$, Eric Sihite$^{2}$, Alireza Ramezani$^{1*}$
\thanks{$^{1}$This author is with the Department of Electrical and Computer Engineering, Northeastern University, Boston MA
		{\tt\small salagame.a, *a.ramezani@northeastern.edu}}%
  \thanks{$^{2}$This author is with California Institute of Technology, Pasadena CA
		{\tt\small esihite@caltech.edu}}
  \thanks{$*$Indicates corresponding author.}
}

\begin{document}

\maketitle
\thispagestyle{empty}
\pagestyle{empty}

\begin{abstract}
Navigating rugged terrain and steep slopes is a challenge for mobile robots. Conventional legged and wheeled systems struggle with these environments due to limited traction and stability. Northeastern University's COBRA (Crater Observing Bio-inspired Rolling Articulator), a novel multi-modal snake-like robot, addresses these issues by combining traditional snake gaits for locomotion on flat and inclined surfaces with a tumbling mode for controlled descent on steep slopes. Through dynamic posture manipulation, COBRA can modulate its heading angle and velocity during tumbling. This paper presents a reduced-order cascade model for COBRA’s tumbling locomotion and validates it against a high-fidelity rigid-body simulation, presenting simulation results that show that the model captures key system dynamics.
\end{abstract}

\section{Introduction}
Rough terrain locomotion remains a challenge for mobile robots due to the irregular surfaces, steep slopes, and varying elevation profiles found in natural environments. Designing effective locomotion strategies for such terrains has been an active area of research \cite{sihite_efficient_2022,sihite_multi-modal_2023}, leading to the development of robots with articulated bodies and deformable structures that can adapt to the terrain. Legged robots, in particular, have shown promise for outdoor locomotion because of their ability to dynamically control ground interactions through intermittent contacts and articulated legs \cite{park_finite-state_2013,hoeller_anymal_2023}. However, achieving reliable locomotion with legged robots on steep slopes is still challenge due to loss of traction \cite{kolvenbach_traversing_2021,ramezani_performance_2013}. These difficulties are further compounded on soft and slippery surfaces, where legs can sink into the ground.

A promising approach for safely descending steep slopes is tumbling structures that exploit gravity to their advantage. This is a concept that has been explored to some degree. NASA's Tumbleweed Rover \cite{behar_nasajpl_2004} for example relied on wind for locomotion. However, in the process it sacrificed controllability for energy efficiency. Other designs such as Spherical Mobile Robot \cite{halme_motion_1996} and University of Pisa's Sphericle \cite{bicchi_introducing_1997} employed an active rolling mechanism with internal weights and a driving wheel. Others such as University of Michigan's Spherobot \cite{mukherjee_simple_1999} and the University of Tehran's August Robot \cite{javadi_a_introducing_2004} use shifting masses inside a spherical shell to generate rolling motion. However, these systems inherently require additional mass for generating inertial forces, leading to increased complexity and weight.

Deformable structures are a more lightweight solution. Platforms such as Ourobot \cite{paskarbeit_ourobotsensorized_2021} and NASA's Tensegrity Robot \cite{bruce_design_2014} utilize body deformation to shift the center of mass for locomotion. Other successful examples are found in \cite{tian_dynamic_2015, wei_design_2019,wang_trajectory_2018, wang_dynamics_2018,sastra_dynamic_2009,sugiyama_crawling_2006}. None of these works have demonstrated dynamic posture control during tumbling, and these solutions are typically slow moving on flat ground, limiting the versatility of these platforms for rugged terrain locomotion.

\begin{figure}
    \centering
    \includegraphics[width=0.75\linewidth]{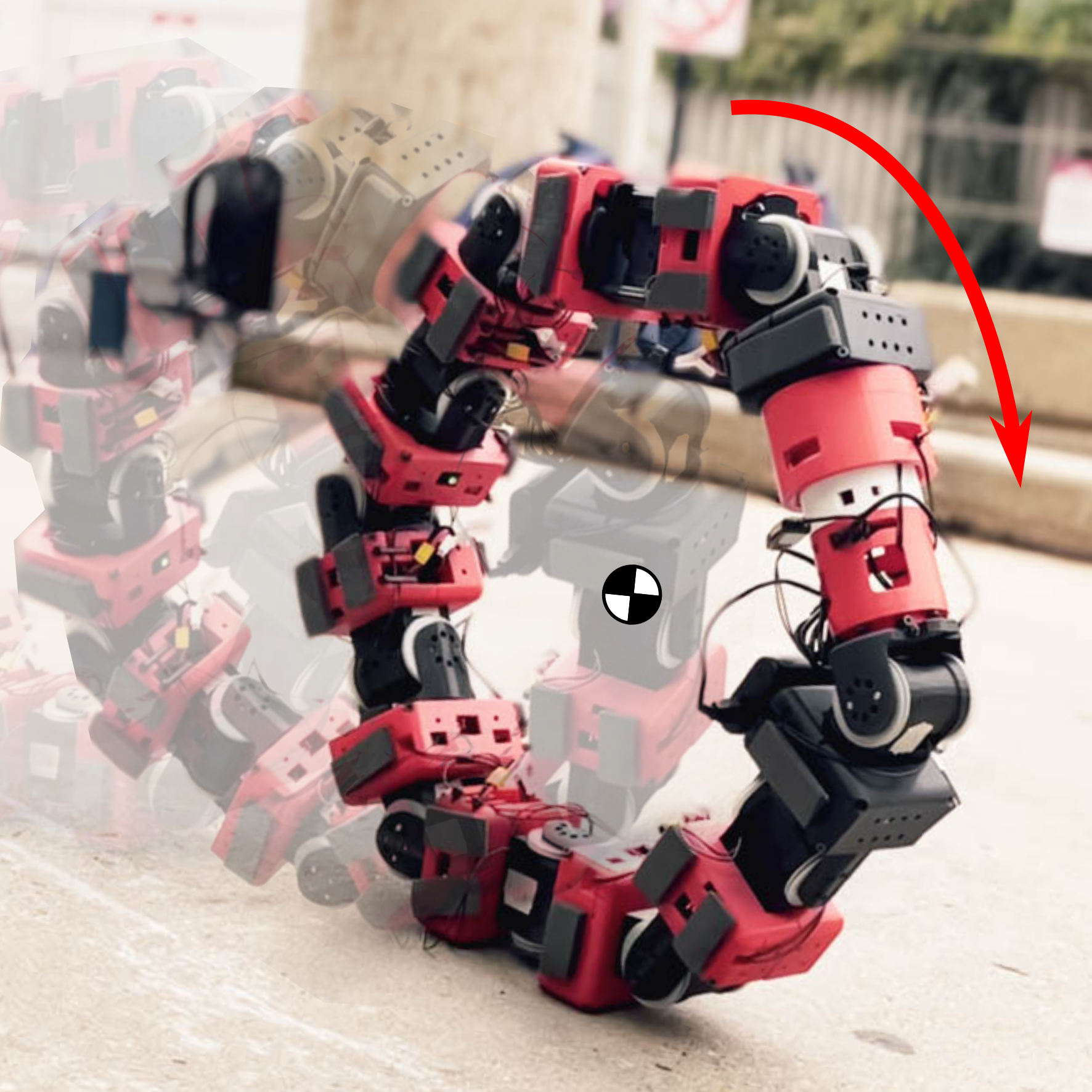}
    \caption{Northeastern University's COBRA robot performing tumbling locomotion}
    \label{fig:cover}
    \vspace{-5mm}
\end{figure}

Our approach to addressing this gap is Northeastern University's COBRA robot. COBRA \cite{salagame_how_2024, salagame_loco-manipulation_2024,salagame_how_2024,salagame_reinforcement_2024,jiang_snake_2024,jiang_hierarchical_2024} is a multi-modal snake inspired robot capable of two distinct types of locomotion: snake-like slithering for navigating flat or uneven terrain, and tumbling for descending steep slopes. In snake mode, COBRA efficiently maneuvers on flat or rugged terrain by using its segmented body to achieve energy-efficient locomotion, distributing its weight across a large contact area to reduce sinking on soft surfaces. Snake gaits such as sidewinding can also be leveraged to travel up steep and slippery slopes \cite{marvi_sidewinding_2014}.

To descend a steep slope, COBRA can transition from a snake configuration into a wheel-like tumbling configuration shown in Fig. \ref{fig:cover} and use its articulated joints to initiate a controlled tumbling motion, exploiting gravity to achieve high-speed descent while maintaining control over its speed and direction through posture adjustments.

Our previous work \cite{salagame_dynamic_2024} introduced a cascaded modeling framework based on a reduced-order model to predict COBRA's behavior during tumbling. This framework modeled the dynamic interactions between posture manipulation inputs and robot states, including heading angle and center of mass velocity. We observed through numerical integration of the cascade model for step and impulse inputs that under certain conditions, there was a deflection in the trajectory of the center of mass of the tumbling structure when an input was applied, indicating that the tumbling trajectory of the system was controllable through this input. 

In this work, we aim to validate this observation by simulating the reduced-order model using a high fidelity physics based simulator. The goal is to make a comparison between the behavior of the simulated reduced-order model and the predicted states from the cascade model and show that it is consistent with previous observations.

The paper is organized as follows: In Section \ref{sec:about}, we briefly describe the COBRA platform, including its hardware and kinematic structure. In Section \ref{sec:rom}, we present our reduced-order modeling framework and cascade model linking posture manipulation inputs and robot states. In Section \ref{sec:validation}, we present our simulation setup to validate the cascade model, and in Section \ref{sec:results} we present the results showing the comparison of the cascade model and the high-fidelity simulation. Finally, we close with some concluding remarks in Section. \ref{sec:conclusion}.

\section{About the COBRA Platform}
\label{sec:about}
\begin{figure}
    \centering
    \includegraphics[width=0.9\linewidth]{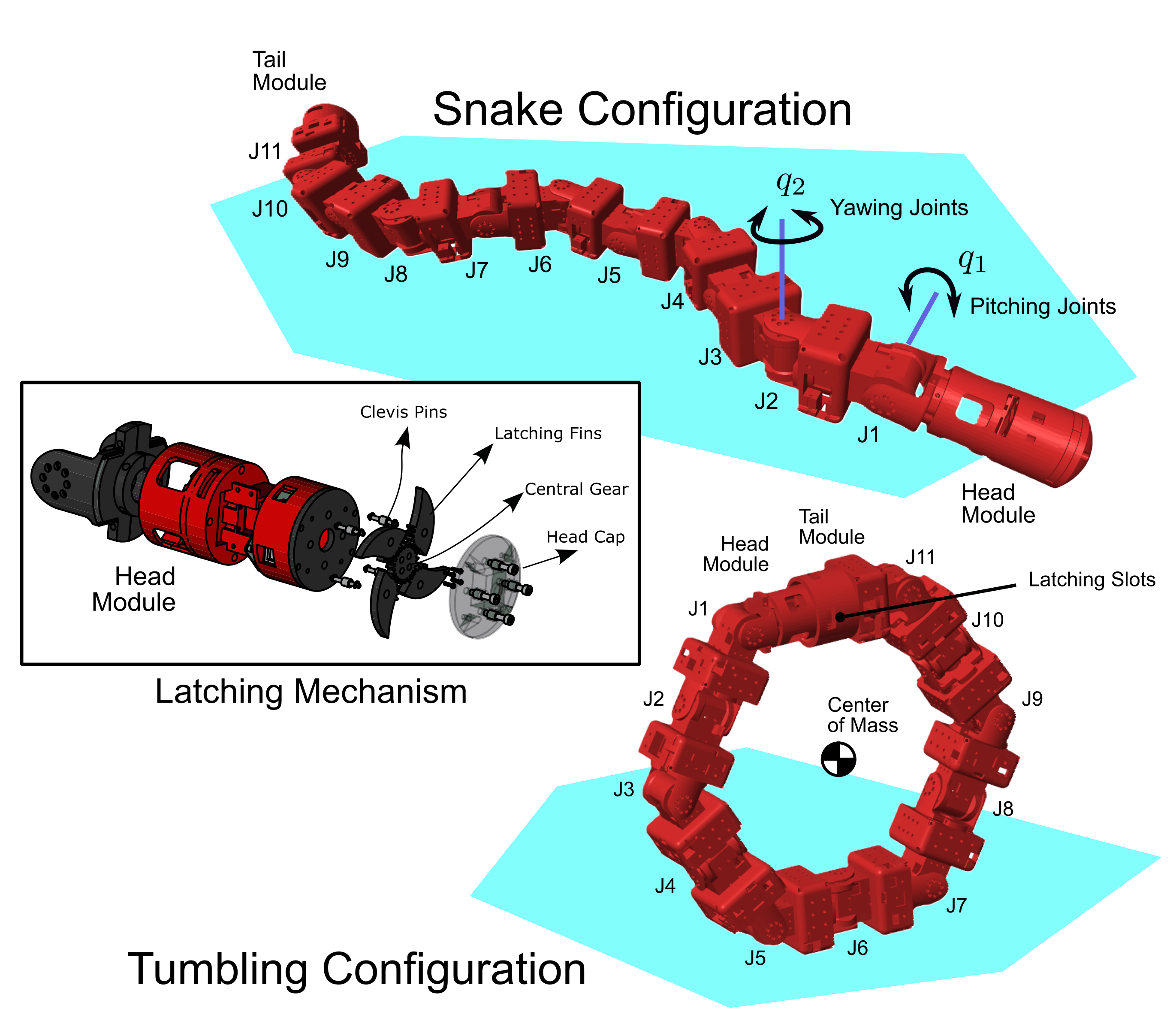}
    \caption{Shows the kinematic structure of COBRA in its snake configuration and tumbling configuration, and latching mechanism used to achieve tumbling confiugration}
    \label{fig:cobra}
    \vspace{-5mm}
\end{figure}

COBRA consists of 11 1-DOF joints alternating between pitching and yawing as shown in Fig. \ref{fig:cobra}. It has 10 identical body links each housing a Dynamixel servo and a battery, and a head module and tail module for a total of 12 links. The head module houses an Nvidia Jetson Orin NX as the main processor, and an Intel RealSense D435i stereo camera equipped with an inertial measurement unit (IMU) for state estimation and navigation. The head module also houses an active latching mechanism that is used to achieve the tumbling configuration shown in Fig. \ref{fig:cobra}. 

COBRA transitions from its snake configuration to its tumbling configuration by raising its head and tail, aligning and locking them using the active latching mechanism integrated into the head module. The latching mechanism enables COBRA's tumbling locomotion \cite{salagame_how_2024} by using four retractable latching fins that remain flush with the surface of the head when closed and deploy via a central gear to engage corresponding slots in the tail module (Fig. \ref{fig:cobra}). This configuration creates a passive rigid connection that withstands forces during tumbling. To initiate tumbling, COBRA transitions into the tumbling configuration at the top of a slope and manipulates its joints to shift its center of mass forward, enabling a controlled descent. Throughout the tumbling process, COBRA can adjust its posture to steer and control the direction of motion.

The following sections outline the dynamic modeling framework for COBRA's tumbling locomotion using a Reduced-Order Model, along with validation through high-fidelity simulations.

\section{Reduced-Order Model (ROM) Derivations}
\label{sec:rom}

\begin{figure}
    \centering
    \includegraphics[width=0.9\linewidth]{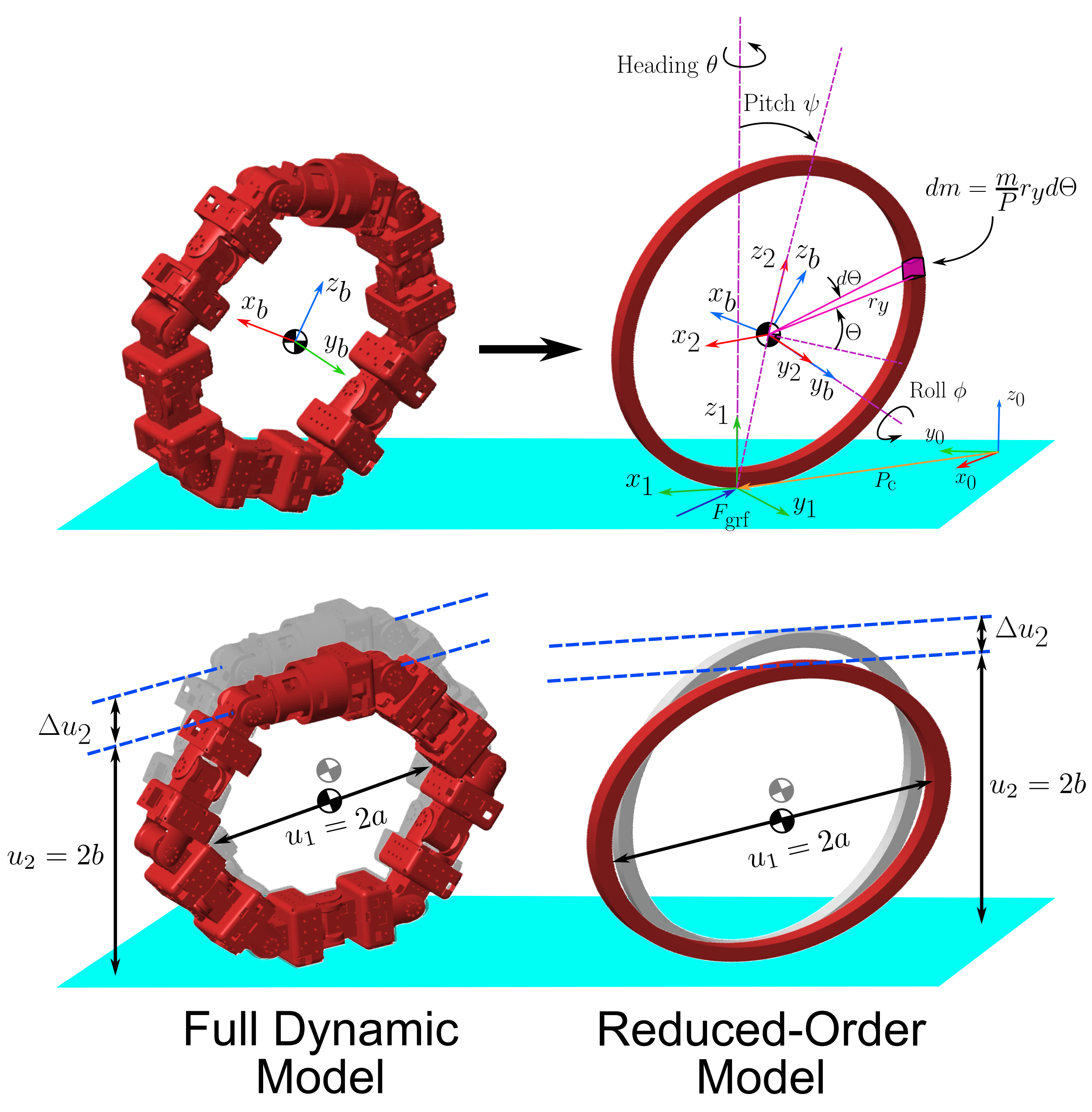}
    \caption{Illustrates the Reduced-Order Model for COBRA's Tumbling Configuration and shows posture manipulation by considering two imaginary actuators, denoted as $u_1$ and $u_2$, which act along the principal axes of the ring to induce planar deformations.}
    \label{fig:rom}
    \vspace{-5mm}
\end{figure}

To simplify the dynamic modeling, we reduce the model of COBRA to a thin elliptical ring as shown in Fig. \ref{fig:rom}. We make the following assumptions about this ring: (1) The ring has a negligible cross-sectional area (Fig. \ref{fig:rom}); (2) The mass is uniformly distributed; (3) The shape of the ring is defined by the principal axes, $u_1$ and $u_2$, controlled by the robot’s joints (Fig.~\ref{fig:rom}); and (4) The postures are symmetric, ensuring the ellipse's center aligns with the center of mass (CoM).

The inertia tensor of this model is then a function of the shape variables $u_i$, given by $\mathcal{I}(u_i)$. Using this tensor, we derive the dynamics equations for tumbling that captures the relation between control actions and tumbling behaviour.

In the following subsections, we briefly describe the dynamic model proposed in our previous work \cite{salagame_dynamic_2024}.

\subsection{Cascade Model}
Our cascaded nonlinear model is structured as follows:
\begin{equation}
    \begin{aligned}
    \Sigma_{tbl}&:
    \begin{aligned}
        \dot x &= f(x,y)
    \end{aligned}
    \\
    \Sigma_{pos}&:\left\{
    \begin{aligned}
        \dot \xi &= f_\xi (\xi,u) \\
        y &= h_\xi(\xi)
    \end{aligned}
    \right.
    \end{aligned}
    \label{eq:cascade-model}
\end{equation}

Here, $x$ and $y$ represent the state vector and output function, which include the ring’s orientation using Euler angles, the CoM position, and the mass moment of inertia about its body axes (x-y-z). The function $f(.)$ governs the state dynamics.

The terms $f_\xi$ and $h_\xi$ represent the dynamics of the internal states $\xi$. The control input $u$ applies actuation along the ring’s principal axes, as shown in Fig. \ref{fig:rom}. The cascade model separates the dynamics into tumbling ($\Sigma_{tbl}$) and posture manipulation ($\Sigma_{pos}$), where tumbling is controlled by internal posture adjustments to maintain the robot's shape in the inertial frame. The following subsections derive the equations for these models.

\subsection{Governing Dynamics for Posture Manipulation $\Sigma_{pos}$}

As shown in Fig.~\ref{fig:rom}, the control actions $u=[u_1,u_2]^\top$ adjust the principal axes for the ring. This section derives the governing equations for posture dynamics.

Consider the general equation of the center line of the elliptical ring in the x-z plane of the body frame with principal axes of length $a$ and $b$,
\begin{equation}
	\frac{p_{i,x}^2}{a^2} + \frac{p_{i,z}^2}{b^2} = 1
	\label{eq:ellipse_equation}
\end{equation}
where $p_i=[p_{i,x},0,p_{i,z}]^\top$ denotes the body-frame coordinates of a point on the ring. Consider the following change of variables:
\begin{equation}
    \begin{aligned}
    & \xi_1=r_y C_\theta, \qquad
    \xi_2=r_y S_\theta
    \end{aligned}
    \label{eq:change-of-variable}
\end{equation}
where $r_y$ and $\theta$ are polar coordinates and are shown in Fig.~\ref{fig:rom}. $S_\theta$ and $C_\theta$ denote $\sin \theta$ and $\cos \theta$. We take the time-derivative of the equation above and Eq.~\ref{eq:ellipse_equation}, which yields 
\begin{equation}
     \begin{aligned}
    & \dot\xi_1=\dot r_y C_\theta-\xi_2\dot\theta, \qquad
    \dot\xi_2=\dot r_y S_\theta+\xi_1\dot\theta\\
    & \frac{\xi_1 \dot\xi_1}{a^2} -\frac{2\xi^2_1 u_1}{a^3}+\frac{\xi_2 \dot\xi_2}{b^2}-\frac{2\xi_2^2u_2}{b^3}=0 \\
    &
    \end{aligned}   
    \label{eq:new-ellipse-eq}
\end{equation}
where $\dot a=u_1$ and $\dot b=u_2$. The perimeter of the ring is fixed and given by the following equation
\begin{equation}
        P = \int^{2\pi}_0 \sqrt{a^2 C^2_{\theta} + b^2 S^2_{\theta}}~ d\theta
\end{equation}
therefore, we can write the following relationship between $\dot P$ and $\dot \theta$
\begin{equation}
        \dot P = \left(\sqrt{a^2 C^2_{\theta} + b^2 S^2_{\theta}}\right)\dot \theta = 0
        \label{eq:perimeter-dot}
\end{equation}
This equation constitutes the remaining ordinary differential equations necessary to establish the state-space model for the posture dynamics. By defining $\xi_3=r_y$, $\xi_4=\theta$, $\xi_5=a$, and $\xi_6=b$, and considering Eqs.~\ref{eq:ellipse_equation}, \ref{eq:change-of-variable}, \ref{eq:new-ellipse-eq}, \ref{eq:perimeter-dot}, the state-space model governing the state vector $\xi=\left[\xi_1,\dots,\xi_6\right]^\top$ is given by
\begin{equation}
    \begin{aligned}
        \begin{bmatrix}
        1 & 0 & -C_{\xi_4} & \xi_2 & 0 & 0\\
        1 & 0 & -S_{\xi_4} & \xi_1 & 0 & 0\\
        \frac{\xi_1}{\xi^2_5} & \frac{\xi_2}{\xi^2_6} & 0 & 0 & 0 & 0\\
        0 & 0 & 0 & \gamma(\xi) & 0 & 0\\
        0 & 0 & 0 & 0 & 1 & 0\\
        0 & 0 & 0 & 0 & 0 & 1\\
        \end{bmatrix}
        \begin{bmatrix}
        \dot \xi_1\\
        \dot \xi_2\\
        \dot \xi_3\\
        \dot \xi_4\\
        \dot \xi_5\\
        \dot \xi_6\\
        \end{bmatrix}&=
        \begin{bmatrix}
        0 & 0\\
        0 & 0\\
        \frac{2\xi_1^2}{\xi_5^3} & \frac{2\xi_2^2}{\xi_6^3}\\
        0 & 0\\
        1 & 0\\
        0 & 1\\
        \end{bmatrix}
        \begin{bmatrix}
        u_1\\
        u_2
        \end{bmatrix}
    \end{aligned}
    \label{eq:pos-dyn-model}
\end{equation}
where $\gamma(\xi)=\sqrt{\xi_5^2 C^2_{\xi_4} + \xi_6^2 S^2_{\xi_4}}$. The matrix in the left-hand side of Eq.~\ref{eq:pos-dyn-model} is invertible, and therefore, the normal form $\dot \xi=f_\xi(\xi,u)$ can be obtained, which is skipped here. Now, it is possible to show that the mass moments of inertia about the body-frame x, y, and z axes, denoted by $\mathcal{I}_{xx}$, $\mathcal{I}_{yy}$, and $\mathcal{I}_{zz}$, are functions of the hidden state vector $\xi$.

The mass of the differential element on the ring can be calculated assuming uniform distribution as follows:
\begin{equation}
	dm = \frac{m}{P} dP = \frac{m}{P} \gamma(\xi) d\xi_4
\end{equation}
where $m$ is the total mass of the elliptical ring. Thus, the mass moment of inertia around each body frame axis can be obtained by:
\begin{equation}
    \begin{aligned}
        \mathcal{I}_{kk} &= \frac{m}{P}\int_{\xi_4} r_k^2\gamma(\xi) d\xi_4,\quad k\in\left\{x, y, z\right\}\\
        r_x&=\xi_3C_{\xi_4}, \quad
        r_y=\xi_3, \quad
        r_z=\xi_3S_{\xi_4}
    \end{aligned}
\end{equation}

In the equation above, the output function $y=h_\xi(\xi)=[\mathcal{I}_{xx},\mathcal{I}_{yy},\mathcal{I}_{zz}]^\top$ encapsulates the mass moments of inertia. Next, we will derive the equations of motion for the tumbling ring using these posture dynamics as follows.

\subsection{Governing Dynamics for Tumbling $\Sigma_{tbl}$}


Consider the ring in Fig.~\ref{fig:rom} equipped with virtual actuators $u_i$ along its principal axes for posture control. We define the following frames of reference: (1) the world frame $x_0$-$y_0$-$z_0$; (2) the contact frame $x_1$-$y_1$-$z_1$ at the contact point $p_c$, with the $z$-axis perpendicular to the ground; (3) the gimbal frame $x_2$-$y_2$-$z_2$ at the CoM $p_{cm}$, inert to the body's motion; and (4) the body frame $x_b$-$y_b$-$z_b$ at the CoM, rotating with the ring.

An inclination is introduced between the contact and world frames, forming an inclined plane at an angle $\alpha$. The ring's orientation $R^0_b$, described in roll, pitch, and yaw angles $\theta$, $\psi$, and $\phi$, is parameterized as follows:
\begin{equation}
    R^0_b = R_z(\theta)R_y(\phi)R_x(\psi)
\end{equation}
The angular velocity vector in the body frame $\omega_b=[\omega_{b,x},\omega_{b,y},\omega_{b,z}]^\top$, in terms of $\dot \theta$, $\dot \psi$, and $\dot \phi$, is expressed as:
\begin{equation}
        \begin{aligned}
        &\omega_{b,x}=\dot{\psi} \sin (\theta) \sin (\phi)+\dot{\theta} \cos (\phi)\\
        &\omega_{b,y}=\dot{\psi} \sin (\theta) \cos (\phi)-\dot{\theta} \sin (\phi)\\
        &\omega_{b,z}=\dot{\psi} \cos (\theta)+\dot{\phi}
        \end{aligned}
	\label{eq:omega_B}
\end{equation}
From the $\Sigma_{pos}$ model, the ring's principal moments of inertia are denoted as $y_1$, $y_2$, and $y_3$. The angular momentum of the ring about $p_{cm}$ is represented by 
\begin{equation}
H_b=
\begin{bmatrix}
    y_1 & 0 & 0\\
    0 & y_2 & 0\\
    0 & 0 & y_3\\
\end{bmatrix}
\omega_b
\label{eq:ang-mom}    
\end{equation}
We define the radius of rotation as the vector extending from $p_{cm}$ to $p_c$. 

Since the ring is in pure rolling at the contact point $p_c$, three constraints must be considered, including one holonomic constraint $(v_{c,z} = 0)$ and two nonholonomic constraints $(v_{c,x} = 0)$ and $(v_{c,y} = 0)$, where $v_c=\left[v_{c,x},v_{c,y},v_{c,z}\right]^\top$ denotes the contact velocity. 

We formulate the equations of motion by resolving the linear and angular momentum balances concerning the ring's CoM. The resulting equations, derived from applying the balance laws alongside the non-integrable constraints, constitute a set of differential equations describing the ring's orientation and the lateral translation of its CoM over time. 

This system of equations is expressed in first-order form 
\begin{equation}
    \dot x = f(x,y)=M^{-1}(x,y)N(x,y)
    \label{eq:tumbling-dyn}
\end{equation}
where the nonlinear terms $M(.)$ and $N(.)$ are given in the Appendix Section. The state vector is represented as $x=[\theta,\psi,\phi,\dot\theta,\dot\psi,\dot\phi,p_{c,x},p_{c,y}]^\top$.

\section{Model Validation in Simulation}
\label{sec:validation}
\begin{figure}
    \centering
    \includegraphics[width=0.9\linewidth]{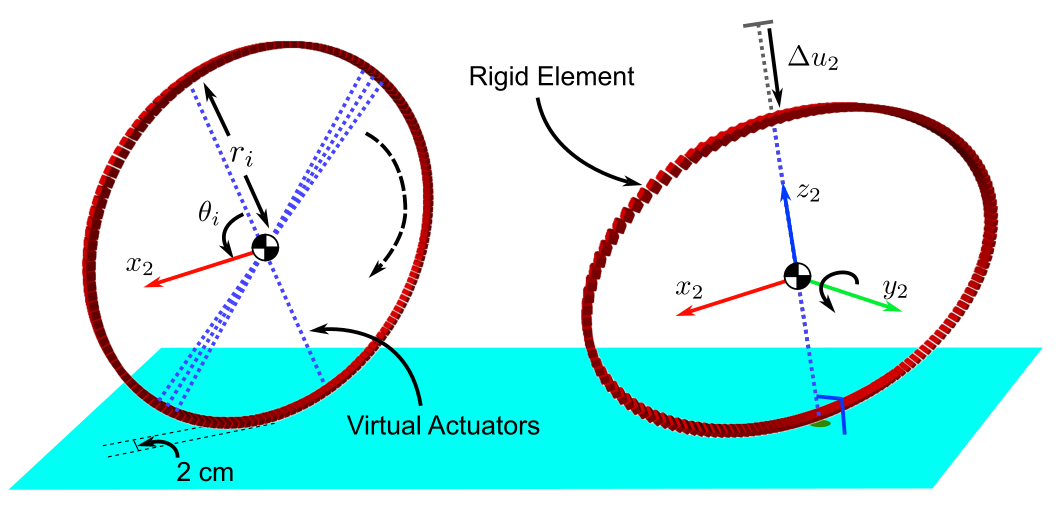}
    \caption{ROM simulation built using Simscape Multi-Body Toolbox to validate presented cascade model dynamics. Shows virtual prismatic actuators used to connect rigid body elements to the center of the ring, approximating a smooth ring.}
    \label{fig:simscape-rom}
    \vspace{-5mm}
\end{figure}
We now describe the simulation setup used to validate the proposed cascade modeling framework. The behavior of the cascade model was simulated using MATLAB's \textit{ode45} to numerically integrate Equation \ref{eq:tumbling-dyn} in response to predefined inputs $u_i(t)$ while tumbling down a 15 deg slope. The resulting states were animated in MATLAB to visualize the system dynamics. 

To evaluate the accuracy of this simulation, we provide the same control input to a high-fidelity simulation of the reduced-order model in MATLAB Simulink using the Simscape Multibody Toolbox, which provides accurate simulation of rigid body dynamics and contact interactions. To represent the smooth, continuous shape of an elliptical ring, a flexible structure is required that can interact rigidly with the ground. To achieve this, we approximate the ring by constructing a structure of 150 discrete rigid elements, as illustrated in Fig. \ref{fig:simscape-rom}. Each element has a mass of $\frac{M}{150}$ and dimensions of $2~\text{cm} \times \frac{P}{150} \times 1~\text{cm}$, where $P$ is the fixed perimeter of the elliptical ring, equal to the full length of COBRA (1.6 m) from head to tail, and $M$ is the total mass of the ring (6 kg, matching COBRA's mass). The rigid elements have a rectangular cross-section to provide stability, while being small enough to uphold the negligible cross-sectional area assumption made in Section \ref{sec:rom}.

\begin{figure}
    \centering
    \includegraphics[width=0.9\linewidth]{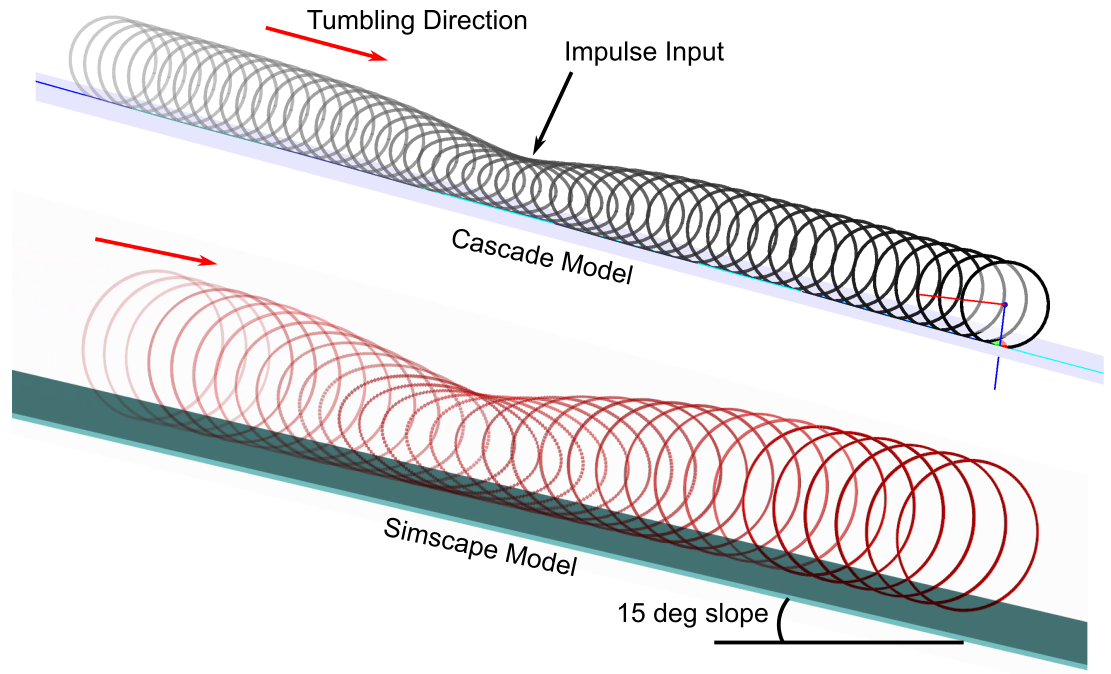}
    \caption{Shows snapshots of Cascade Model (above) and Simscape Model (below) executing impulse input during tumbling}
    \label{fig:snapshots}
\end{figure}
\begin{figure}
    \centering
    \includegraphics[width=0.9\linewidth]{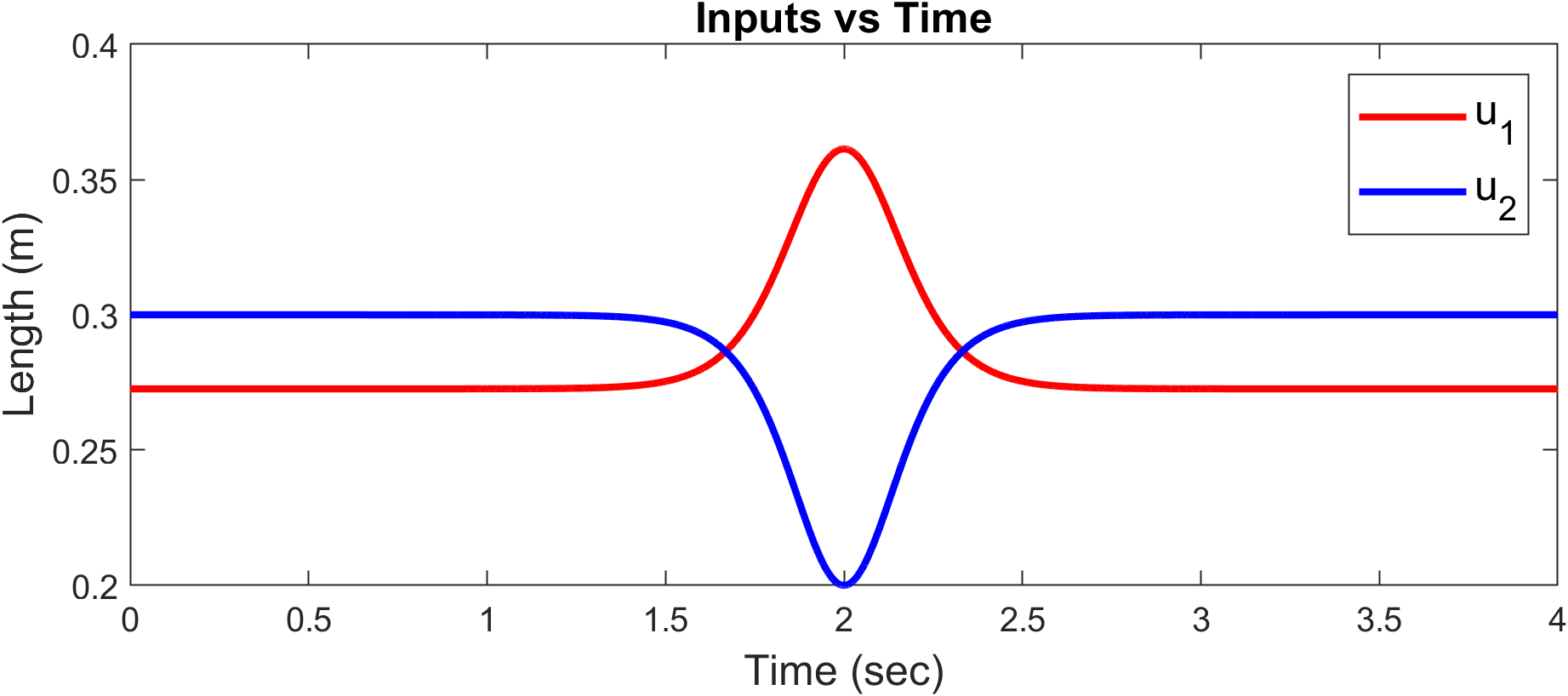}
    \caption{Shows the input signal provided to the models. Due to the constraint of fixed perimeter according to equation \ref{eq:ellipse_equation}, as $u_1$ increases, $u_2$ increases.}
    \label{fig:input}
    \vspace{-5mm}
\end{figure}

Each rigid element is connected to the center of the ring using prismatic actuators that control the distance $r_i$ between the $i^\text{th}$ element and the ring center. The actuation is mirrored such that the diametrically opposite element is always positioned at the same distance $r_i$, upholding the symmetricity assumption made in Section \ref{sec:rom}. These elements are arranged at regular intervals of $2\pi/150$ radians around the ring to complete the structure. To preserve the elliptical shape and apply the inputs $u_i = [a,~b]$ of the cascade model, the following mapping is derived by rearranging equation \ref{eq:ellipse_equation}:

\begin{equation}
    r_i = \frac{ab}{(b^2C_{\theta_i}+a^2S_{\theta_i})^\frac{1}{2}}
    \label{eq:sim-actuator}
\end{equation}

where $\theta_i$ is measured with respect to the non-rotating body frame $[x_2,~y_2,z_2]$ as defined in Fig. \ref{fig:rom}. Using Equation \ref{eq:sim-actuator}, we can smoothly deform the shape of the ring given parameters $[a,~b]$ as shown in Fig. \ref{fig:simscape-rom}.

The contact dynamics between the rigid elements and the ground are defined by the Spatial Contact Force Block in Simscape, that uses a smooth spring damper model:

\begin{equation}
f_n = s(d, w) \cdot (k \cdot d + b \cdot \dot d~),
\end{equation}

where $f_n$ is the normal force, $d$ is the penetration depth, $w$ is the transition region width, $k$ and $b$ are the spring stiffness and damping coefficient, and $s(d,w)$ is a smoothing function. For this simulation, a transition region of $10^{-3}$m, spring stiffness of $10^{4}$N/m and damping coefficient $10^3$N/ms are used. For friction force, the model uses a smooth stick-slip model defined by:
\begin{equation}
    |f_f| = \mu\cdot |f_n|
\end{equation}
where $f_f$ is the friction force and $\mu$ is the friction coefficient. The direction of $|f_f|$ is always opposed to the direction of relative velocity between the two surfaces. To simulate close to no-slip conditions, a high coefficient of friction of $5$ is used.

In the following section, we present a comparison between the behavior of the reduced-order model as predicted by integrating the cascade model, and the behavior of the high fidelity simulation in MATLAB Simulink.

\section{Results}
\label{sec:results}
Figure \ref{fig:snapshots}  illustrates snapshots of the reduced-order model (ROM) tumbling as predicted by both the cascade model and the high-fidelity Simscape model. The input signal provided to each model is depicted in Figure \ref{fig:input}. This signal takes the form:
$$
b(t) = 4b^\prime\sigma(\gamma(t-t_0))(1-\sigma(\gamma(t-t_0))) + b_0
$$ 
where $\sigma$ is the Sigmoid function parameterized by variables $b^\prime$, $t_0$ and $\gamma$ representing the amplitude, time and sharpness of the impulse peak respectively. Here, $b_0$ denotes the initial length of the axis $b$ under zero input. For the simulation, we use parameters $\gamma$ of 10, $t_0$ of 2, and $b_0$ and $b^\prime$ of 0.3m and 0.2m respectively.

Based on our prior work \cite{salagame_dynamic_2024}, it was observed that applying control inputs when the ROM exhibits non-zero angular velocity about all axes leads to changes in the heading angle. To validate this, both models were initialized with a tumbling velocity $\dot{\phi} = 2\pi$ rad/s and a roll angular rate $\dot{\psi} = \pi/6$ rad/s. The simulation was run for 4 seconds, with the impulse input peaking at the 2-second mark.

\begin{figure}[t]
    \centering
    \includegraphics[width=0.9\linewidth]{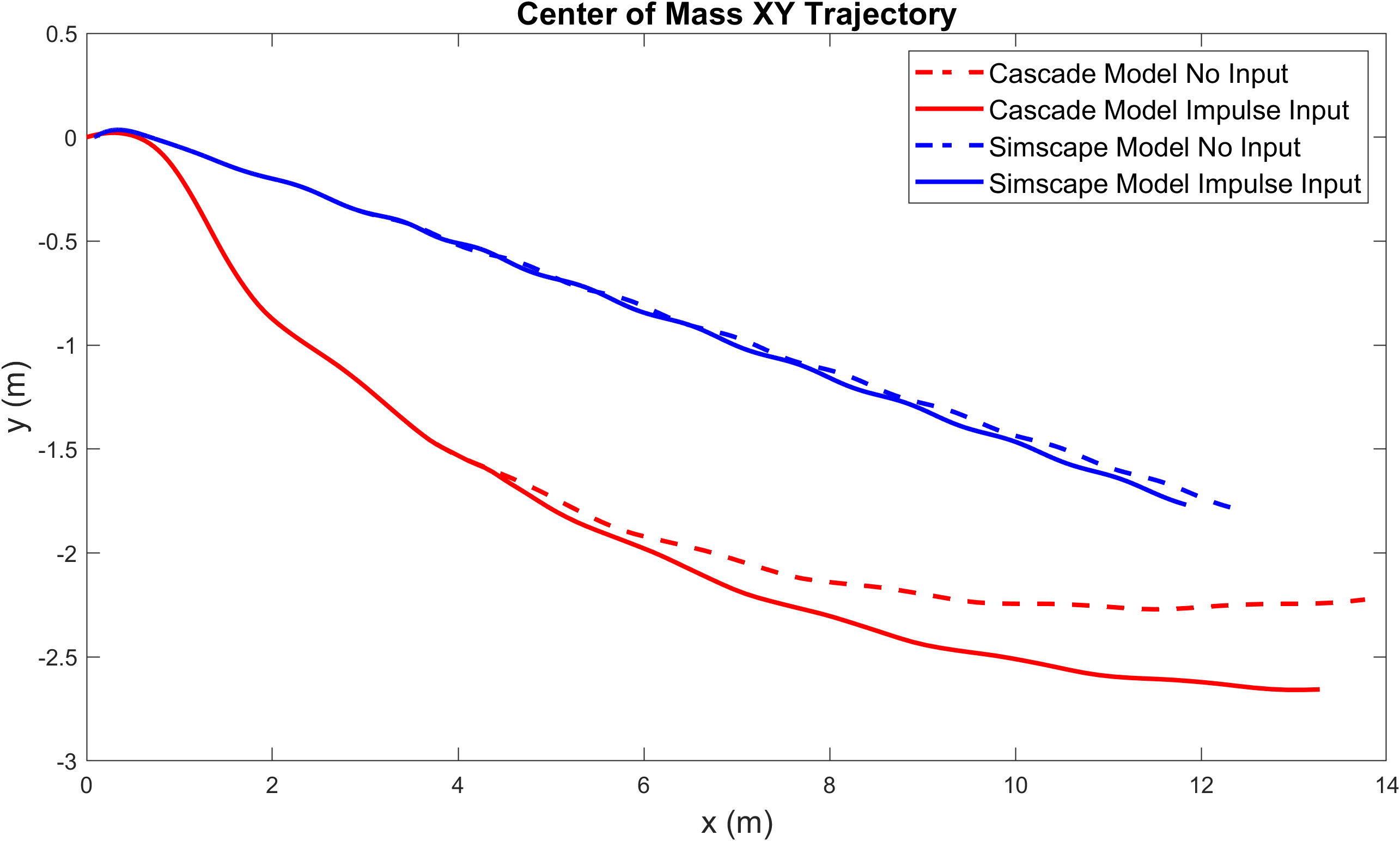}
    \caption{Shows the trajectory of the Center of Mass during tumbling for both models with and without application of input}
    \label{fig:trajectory}
\end{figure}

\begin{figure}[t]
    \centering
    \includegraphics[width=0.9\linewidth]{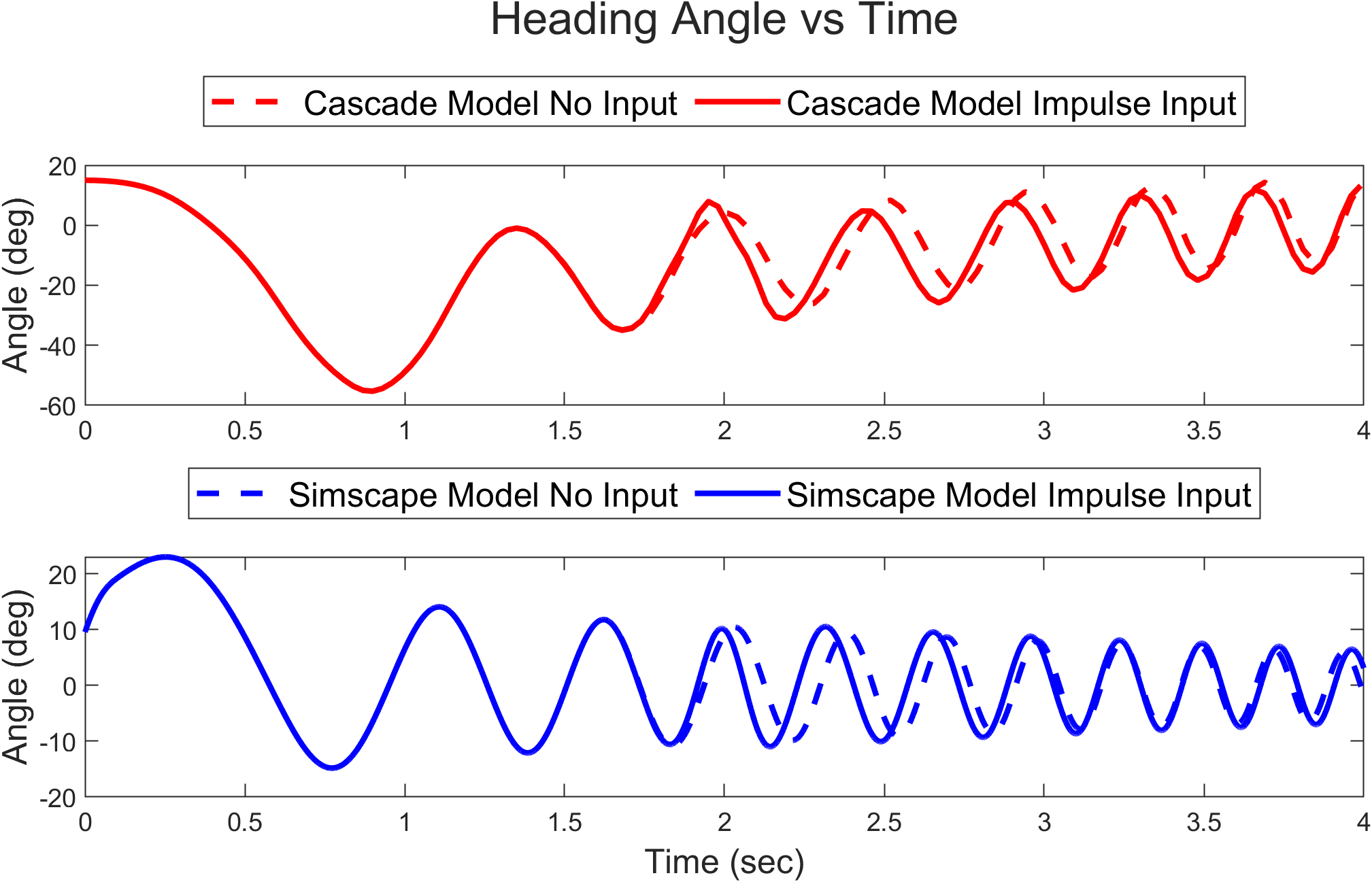}
    \caption{Comparison of heading angle vs time for both models}
    \label{fig:heading}
    \vspace{-5mm}
\end{figure}
    
\begin{figure}[t]
    \centering
    \includegraphics[width=0.9\linewidth]{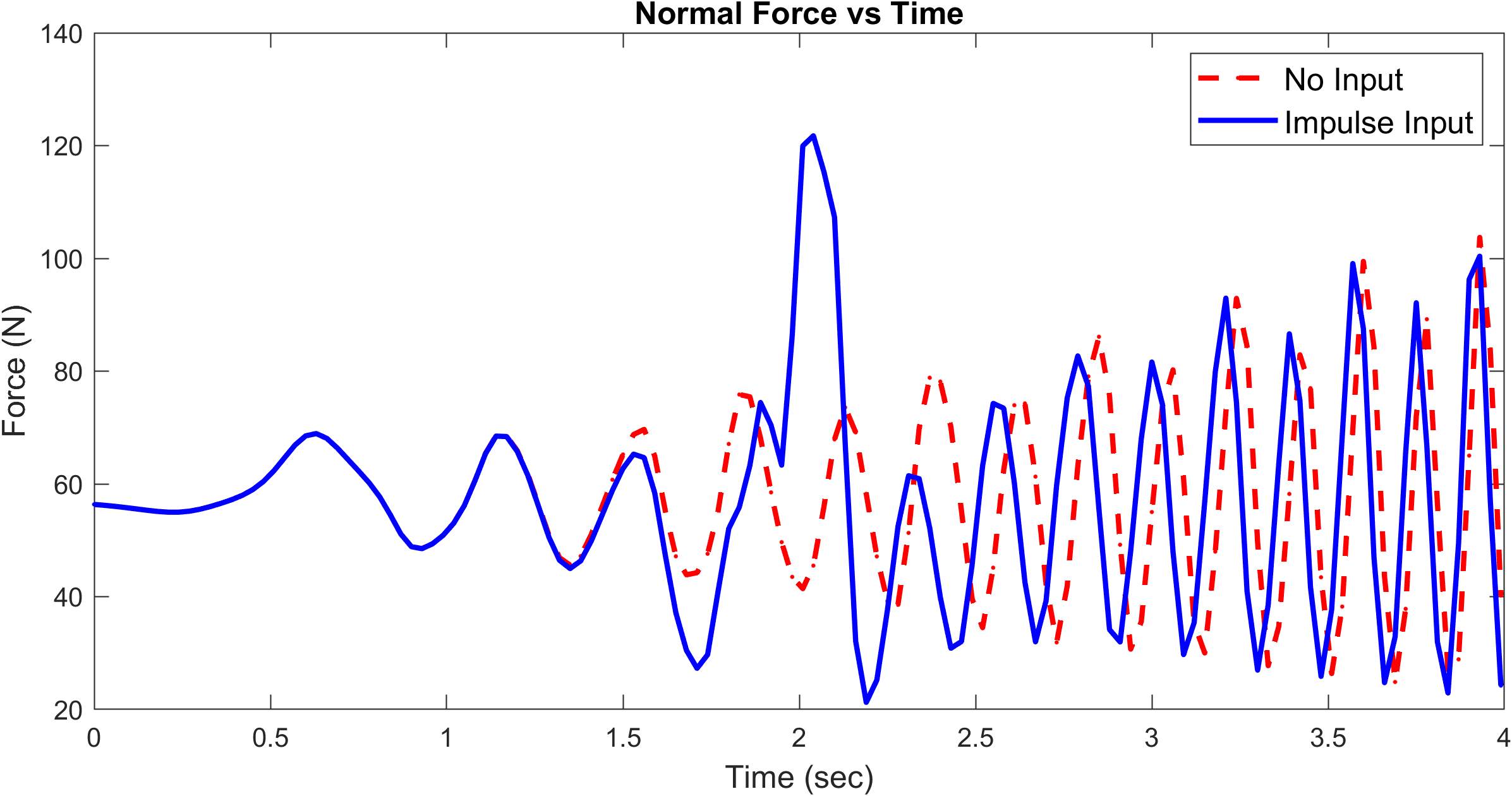}
    \caption{Normal forces predicted by Cascade Model}
    \label{fig:grf}
\end{figure}

Figure \ref{fig:trajectory} shows the trajectory of the center of mass (CoM) during tumbling for both the cascade and high-fidelity Simscape models, under both input and no-input conditions. The cascade model operates under idealized assumptions regarding slippage at the contact points and inertia distribution of the ring, leading to a more simplified dynamic response. In contrast, the Simscape model captures a higher degree of realism, including more accurate inertia for a less idealized mass distribution, and contact dynamics. Additionally, Simscape treats initial conditions as a best-effort approximation, resulting in slight discrepancies at the start of the motion compared to the cascade model. These combined factors contribute to the differences observed in the CoM trajectory. Nonetheless, after applying the control input, both models exhibit a similar directional response, with the Simscape model showing reduced deviations due to the more realistic dynamic interactions captured in its framework.

Figure \ref{fig:heading}  compares the heading angle evolution over time for both models, showing a strong agreement overall. A critical assumption in the cascade model is that the ring maintains continuous ground contact with a non-zero positive normal force throughout its motion. However, this constraint is not explicitly enforced in Simscape, leading to notable differences when the cascade model predicts negative normal forces. In such cases, while the cascade model artificially preserves ground contact, the Simscape model accurately transitions to ballistic motion, reflecting a more realistic loss of contact. To prevent this undesired behavior, we carefully select input signals that ensure positive ground reaction forces throughout the simulation. Figure \ref{fig:grf} shows the ground reaction forces for the cascade model under the applied input signal. 

\begin{figure}
    \centering
    \includegraphics[width=0.9\linewidth]{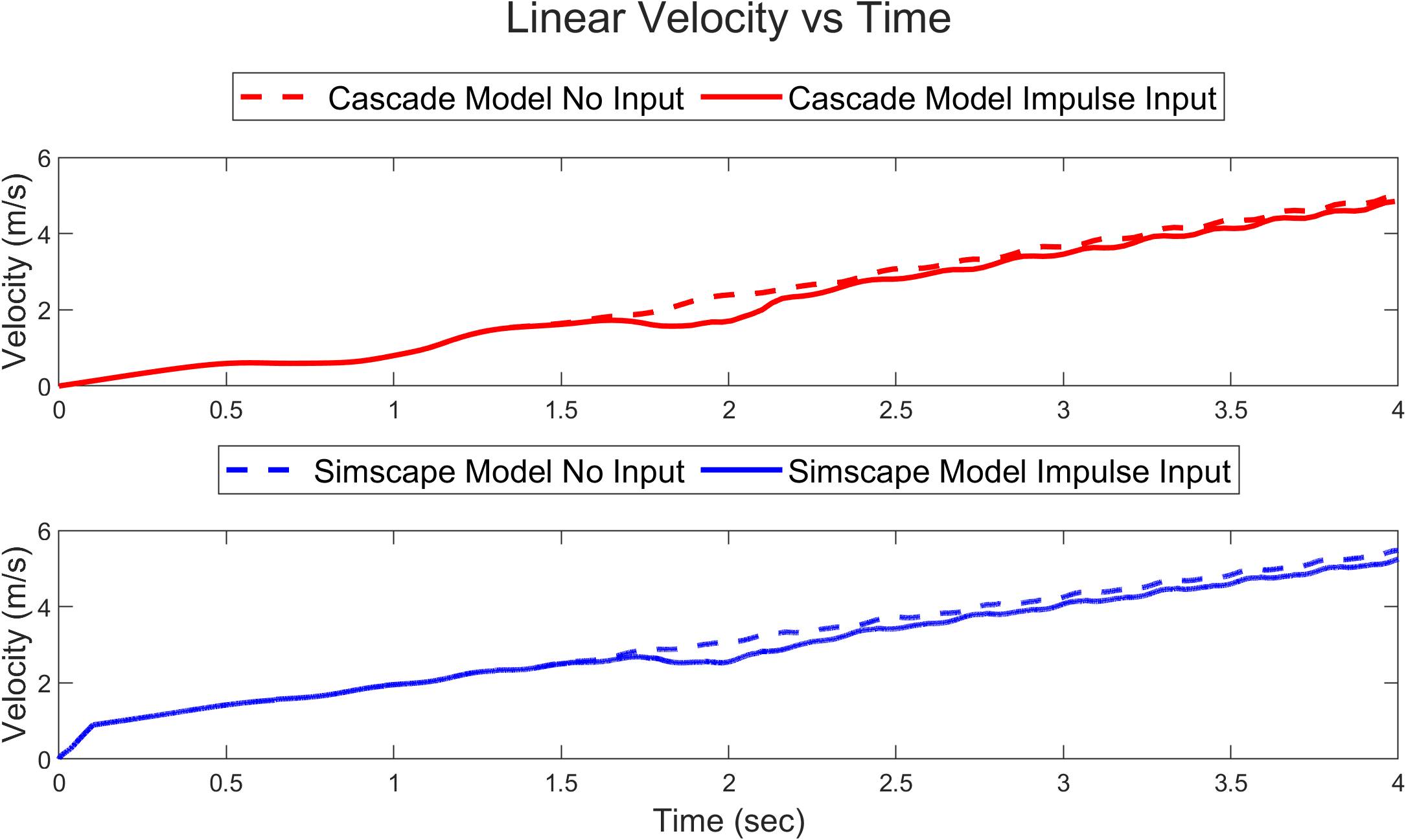}
    \caption{Velocity of Center of Mass in the direction of tumbling}
    \label{fig:velocity}
\end{figure}
Finally, Figure \ref{fig:velocity} depicts the evolution of the CoM's linear velocity during tumbling for both models. The variations in velocity stem from changes in inertia induced by control actions. Once again, the two models exhibit strong alignment, highlighting the fidelity of the reduced-order cascade model in capturing the system's dynamics.

\section{Conclusion}
\label{sec:conclusion}
The objective of this work was to validate the proposed modeling framework and assess its capability in predicting the behavior of a tumbling structure undergoing posture manipulation. While these results are presented for a single type of input signal, this specific signal was selected as a representative case as variations of it consistently produced a change in heading angle and velocity during tumbling, as shown in prior studies. The results indicate that the cascade model closely approximates the high-fidelity Simscape simulation in capturing the system’s dynamics under external inputs. Although some deviations arise due to unmodeled effects like slippage, the model accurately reflects key behaviors, such as changes in the center of mass trajectory, heading angle, and linear velocity, demonstrating good qualitative agreement overall. This supports the validity of the cascade model’s assumptions and establishes its suitability for integration into a closed-loop controller for trajectory tracking during tumbling. 
Future efforts will leverage this validated model to develop a controller for the reduced-order Simscape representation, ultimately extending to full-scale control of COBRA’s dynamics.

\balance{}
\printbibliography
\end{document}